\documentclass{article}

\usepackage{PRIMEarxiv}

\usepackage[utf8]{inputenc} % allow utf-8 input
\usepackage[T1]{fontenc}    % use 8-bit T1 fonts
\usepackage{hyperref}       % hyperlinks
\usepackage{url}            % simple URL typesetting
\usepackage{booktabs}       % professional-quality tables
\usepackage{amsfonts}       % blackboard math symbols
\usepackage{nicefrac}       % compact symbols for 1/2, etc.
\usepackage{microtype}      % microtypography
\usepackage{fancyhdr}       % header
\usepackage{graphicx}       % graphics
\graphicspath{{media/}}     % organize your images and other figures under media/ folder
\usepackage{amsmath}        % for math formatting
\usepackage{amssymb}        % for mathematical symbols

\title{PyScrew: A Comprehensive Dataset Collection from Industrial Screw Driving Experiments}

\author{
  Nikolai West \\
  RIF Institute for Research and Transfer e.V. \\
  Work and Production systems \\
  Dortmund, Germany \\
  \\
  Institute for Production Systems \\
  Technical University Dortmund \\
  Dortmund, Germany \\
  \texttt{nikolai.west@tu-dortmund.de} \\
   \And
  Jochen Deuse \\
  Institute for Production Systems \\
  Technical University Dortmund \\
  Dortmund, Germany \\
  \\
  Centre for Advanced Manufacturing \\
  University of Technology Sydney \\
  Sydney, Australia \\ 
  \texttt{jochen.deuse@tu-dortmund.de} \\
}

\begin{document}
\maketitle

\begin{abstract}
This paper presents a comprehensive collection of industrial screw driving datasets designed to advance research in manufacturing process monitoring and quality control. The collection comprises six distinct datasets with over 34,000 individual screw driving operations conducted under controlled experimental conditions, capturing the multifaceted nature of screw driving processes in plastic components. Each dataset systematically investigates specific aspects: natural thread degradation patterns through repeated use (s01), variations in surface friction conditions including contamination and surface treatments (s02), diverse assembly faults with up to 27 error types (s03-s04), and fabrication parameter variations in both upper and lower workpieces through modified injection molding settings (s05-s06). We detail the standardized experimental setup used across all datasets, including hardware specifications, process phases, and data acquisition methods. The hierarchical data model preserves the temporal and operational structure of screw driving processes, facilitating both exploratory analysis and the development of machine learning models. To maximize accessibility, we provide dual access pathways: raw data through Zenodo with a persistent DOI, and a purpose-built Python library (PyScrew) that offers consistent interfaces for data loading, preprocessing, and integration with common analysis workflows. These datasets serve diverse research applications including anomaly detection, predictive maintenance, quality control system development, feature extraction methodology evaluation, and classification of specific error conditions. By addressing the scarcity of standardized, comprehensive datasets in industrial manufacturing, this collection enables reproducible research and fair comparison of analytical approaches in an area of growing importance for industrial automation.
\end{abstract}


\section{Introduction}
\label{sec:introduction}
Assembly processes in manufacturing settings generate rich multivariate time series data that can reveal critical insights about product quality and process stability. Among these, screw driving operations are particularly ubiquitous, yet the industrial research community faces a significant obstacle: the scarcity of standardized, comprehensive datasets for developing and validating data-driven methodologies. This obstacle limits progress in applying advanced anomaly detection methods \cite{schlegl2021a}, feature extraction techniques \cite{west2021}, and classification algorithms \cite{schlegl2021b} to industrial time series data. While proprietary datasets exist within industrial environments, their restricted access creates barriers to reproducible research and fair comparison of analytical approaches.

This paper introduces PyScrew, a comprehensive open collection of six screw driving datasets with over 34,000 individual operations collected under controlled experimental conditions \cite{zenodo2025}. Building upon the pioneering work of Leporowski et al. \cite{leporowski2021} with the \href{https://zenodo.org/records/4487073}{AURSAD} dataset, PyScrew expands both the scale and diversity of available industrial time series data. We organize these experimental conditions as scenarios \textbf{s} with a simple numbering system for clear identification. These datasets systematically capture various aspects of the screw driving process: natural thread degradation patterns \textbf{(s01)} \cite{west2024a}, workpiece surface condition variations \textbf{(s02)} \cite{west2024b, henkies2025}, component faults with multiple error types \textbf{(s03-s04)} \cite{west2025}, and injection molding parameter variations \textbf{(s05-s06)}. Each dataset is fully documented with detailed experimental protocols to ensure transparency and proper interpretation.

The datasets follow a hierarchical data model that preserves the structure of screw driving operations. This design facilitates both exploratory analysis and the development of machine learning models for anomaly detection \cite{west2024a}, process monitoring, and fault diagnosis in industrial settings. The PyScrew data is made available through \href{https://doi.org/10.5281/zenodo.14729547}{Zenodo}, an online storage service for scientific data sets, and can be easily accessed either through GitHub (\url{https://github.com/nikolaiwest/pyscrew}) or PyPi (\url{https://pypi.org/project/pyscrew/}), where we provide a custom library designed for working with these datasets in Python.

These datasets are designed for various research applications, including:
\begin{itemize}
    \item Development of machine learning models for anomaly detection and classification of specific error types \cite{west2025}
    \item Process monitoring and quality control system development
    \item Manufacturing analytics and parameter optimization
    \item Digital twin development for screw driving operations
    \item Analysis of material property influences on assembly processes
    \item Evaluation of both supervised and unsupervised approaches to quality control \cite{west2024a}
\end{itemize}

The remainder of this paper is organized as follows: Section~\ref{sec:related_work} (Related Work) reviews relevant literature in the field of industrial datasets and screw driving analysis. Section~\ref{sec:experimental_setup} (Experimental Setup) describes the hardware components, process phases, and data acquisition methods used to collect the data. Section~\ref{sec:dataset_description} (Dataset Description) details each of the six datasets, including their specific experimental conditions and characteristics. Section~\ref{sec:data_access} (Data Access) explains how to access the data, both directly through Zenodo and using the PyScrew library. Section~\ref{sec:limitations} (Limitations and Future Work) discusses constraints of the current collection and potential extensions, and Section~\ref{sec:conclusion} (Conclusion) summarizes the contributions and implications of this work.

\section{Related Work}
\label{sec:related_work}

This section reviews some literature in the field of industrial manufacturing data analysis with a focus on screw driving operations. While considerable research has focused on developing methodologies for analyzing manufacturing time series, including techniques for anomaly detection and feature extraction, the availability of publicly accessible, comprehensive datasets for screw driving operations remains severely limited. The AURSAD dataset by Leporowski et al. \cite{leporowski2021} stands as a notable exception prior to our contribution. Given this significant data scarcity in the domain and our group's sustained research efforts, this section primarily focuses on the methodological works most pertinent to screw driving process analysis, among which a significant portion originates from our previous studies. It is important to note that this section provides a focused overview of this specific area, rather than a fully exhaustive review of all related literature. A comprehensive systematic literature review of machine learning methods in this domain, offering a broader perspective, is currently underway and is intended for future publication.

The remainder of this section is structured to provide a detailed overview of the relevant landscape. We begin by examining the availability of public datasets in industrial manufacturing, highlighting the current limitations and gaps that our work addresses. Next, we explore research on anomaly detection in manufacturing time series. We then review feature extraction and classification techniques specifically developed for industrial time series data, with particular attention to methods for imbalanced datasets. The section continues with an overview of process monitoring approaches for screw driving operations. Finally, we discuss existing frameworks for time series analysis in manufacturing and position our contribution within this landscape. Throughout this review, we highlight how the PyScrew dataset collection addresses limitations in current research and provides opportunities for advancing the state of the art in manufacturing process monitoring.

\subsection{Public Datasets in Industrial Manufacturing} 

Industrial manufacturing data, particularly for assembly processes like screw driving, remains limited in public repositories compared to other domains. This scarcity creates significant barriers to reproducible research and fair comparison of analytical approaches. Leporowski et al. \cite{leporowski2021} made a pioneering contribution with the AURSAD (Aarhus University Robot Screwdriving Anomaly Detection) dataset, which was the first open-source dataset specifically designed for screwdriving anomaly detection research. This dataset contains 2,045 samples (1,420 normal and 625 anomalous) with 134 features collected using a Universal Robot UR3e and an OnRobot Screwdriver. AURSAD includes time-series data for five types of operations: normal operations, missing screw anomalies, damaged screw thread anomalies, extra assembly component anomalies, and damaged plate thread anomalies. The authors also provided a companion Python library to facilitate data processing.

While \href{https://github.com/CptPirx/AURSAD}{AURSAD} provided an important starting point for open research in this area, it is limited in both scale (approximately 2,000 samples) and scope (focusing on only five operation types). The PyScrew collection presented in this paper significantly expands upon this foundation by providing six distinct datasets with over 34,000 screw driving operations that capture a much wider range of experimental conditions, anomaly types, and process variations. Unlike AURSAD and many existing industrial datasets that focus on a single aspect of a process, this collection provides comprehensive coverage of different factors affecting screw driving operations, from material properties to assembly conditions and specific error types.

Most research on screw driving processes in industrial settings has relied on proprietary datasets that are not publicly accessible. This creates challenges for reproducibility and makes it difficult to compare the performance of different analysis approaches fairly. By providing open access to well-documented screw driving datasets through a persistent DOI \cite{zenodo2025}, this work aims to advance research in industrial quality control and process monitoring by enabling reproducible research and fair comparisons between different methodologies.

\subsection{Anomaly Detection in Manufacturing Time Series}

The detection of anomalies in manufacturing time series data represents a critical application area for quality control systems. West and Deuse \cite{west2024a} conducted a comparative study of machine learning approaches for anomaly detection in industrial screw driving data (s01), highlighting the need for standardized datasets in this field. Their work demonstrated that both supervised models like Random Forest (achieving 99.02\% accuracy and 98.36\% F1-score) and unsupervised approaches like DBSCAN (achieving 96.68\% accuracy and 90.70\% F1-score) can effectively detect anomalies in screw driving operations.

In the manufacturing domain, interpretable models for anomaly detection are particularly valuable. Schlegl et al. \cite{schlegl2021a} addressed this challenge by developing an interpretable deep learning approach for anomaly detection in manufacturing systems that both improves model transparency and enhances scalability, a key requirement for industrial deployment. Their work demonstrated that learning interpretable shapes from time series data not only facilitates human understanding of model decisions but also improves anomaly detection performance when applied to screw driving operations in automotive assembly.

The challenge of working with imbalanced datasets is particularly relevant in manufacturing, where normal operations vastly outnumber anomalies. West et al. \cite{west2021} addressed this challenge for their k-means clustering work by implementing Dynamic Time Warping (DTW) as a distance metric specifically suited for screw driving time series data, which enabled effective unsupervised anomaly detection in imbalanced datasets consisting of 50,000 normal and 96 anomalous samples. This approach demonstrates the importance of domain-specific distance metrics when working with industrial time series data.

Addressing the challenge of setting effective thresholds for fault detection, Schlegl et al. \cite{schlegl2022a} proposed an algorithm for the automated search of process control limits in time series data. Conceptualized to mimic the systematic steps taken by domain experts, this approach automates the typically manual process of defining control limits, demonstrating efficacy on real-world manufacturing data and achieving strong performance on benchmark datasets.

\subsection{Feature Extraction and Classification for Industrial Time Series}

For time series analysis in manufacturing settings, West et al. \cite{west2021} proposed another  approach using DTW for feature extraction that significantly reduces computational effort while maintaining competitive classification performance. Their method extracts maximally discriminative features from multivariate time series data, achieving comparable results to more computationally intensive approaches on real-world manufacturing data with a 97.90\% reduction in computational effort. In the broader context of time series classification in manufacturing, specialized feature extraction approaches like those proposed by West et al. \cite{west2021} have shown promise for reducing dimensionality while preserving discriminative information.

Building on this work, Henkies et al. \cite{henkies2025} conducted a comprehensive evaluation of feature extraction methods for screw connection data, focusing specifically on the s02 dataset with surface-based anomalies. Their comparative analysis of PAA, PCA, catch22, and tsfresh extraction methods demonstrated that different approaches have distinct advantages depending on whether the priority is computational efficiency, memory requirements, or classification performance. This work demonstrated that not only can anomalies be detected, but specific error types can be classified with high accuracy using time series features from screw driving data.

For classifications tasks with imbalanced industrial data, Schlegl et al. \cite{schlegl2021b} proposed a margin-based greedy shapelet search algorithm specifically designed for robust classification of imbalanced time series data, demonstrating superior performance compared to traditional approaches when applied to manufacturing process data. Their approach focuses on identifying discriminative subsequences (shapelets) that maximize the margin between different classes, which is particularly valuable when working with rare fault conditions that may have limited examples in the training data.

\subsection{Process Monitoring for Screw Driving Operations}

Most recently, West and Deuse \cite{west2025} expanded the scope of screw driving process monitoring by developing a multi-class error detection framework capable of distinguishing between 25 distinct error types in industrial screw driving operations (s04 dataset). Their findings revealed varying detectability across different error categories, with component/thread modifications and environmental conditions being particularly distinguishable. This work demonstrated that process monitoring systems can go beyond binary anomaly detection to provide specific diagnostic information about the nature of detected faults.

Their subsequent work on detecting surface-based anomalies for self-tapping screws in plastic housings \cite{west2024b} (s02) further demonstrated the value of comprehensive time series data for quality control applications. This study focused specifically on how different surface conditions affect the screw driving process, showing that contamination, wear, and surface treatments each produce characteristic signatures in the torque and angle profiles that can be detected using machine learning approaches.

These studies highlight the complexity of screw driving processes and the need for detailed data that captures the various factors affecting connection quality. Unlike simpler assembly operations, screw driving involves multiple phases (finding, driving in, pre-tightening, and final tightening) that each provide different diagnostic information about the process. Comprehensive datasets that capture the full process with high temporal resolution are therefore particularly valuable for developing robust monitoring systems.

\subsection{Frameworks for Time Series Analysis in Manufacturing}

Several frameworks exist for general-purpose time series analysis, such as \href{https://github.com/cesium-ml/cesium}{Cesium} or \href{https://github.com/tslearn-team/tslearn}{tslearn} \cite{tavenard2020}, but these lack domain-specific functionality for industrial manufacturing data. General-purpose frameworks typically focus on features and distance metrics that may not be optimal for the specific characteristics of manufacturing time series, such as the multi-phase nature of assembly operations or the importance of gradient features in screw driving.

Domain-specific implementations have shown advantages for manufacturing applications. West et al. \cite{west2021} demonstrated this by implementing Dynamic Time Warping specifically tuned for screw driving data, achieving better performance than generic distance metrics. Similarly, the feature extraction evaluation by Henkies et al. \cite{henkies2025} showed that domain knowledge about the screw driving process can inform the selection of appropriate dimensionality reduction techniques, improving both computational efficiency and classification performance.

While general frameworks exist for time series analysis, tools for effectively retrieving specific patterns from massive industrial time series databases are also critical. Addressing this, Schlegl et al. \cite{schlegl2022b} presented a novel adaptive similarity search algorithm designed to improve the recall of information retrieval systems for large manufacturing time series databases. This approach supports domain experts in iteratively querying millions of parts to identify specific process faults, utilizing relevance feedback and self-adaptation to refine searches and build a library of fault patterns, showing improved performance on real-world manufacturing and benchmark data.

The PyScrew library presented in this paper aims to bridge this gap by providing a specialized framework for working with screw driving datasets. By combining standardized data access with domain-specific preprocessing options, PyScrew enables researchers to focus on developing and testing analytical methods rather than data preparation. This approach follows the example set by the AURSAD library but expands both the scale of available data and the flexibility of the processing pipeline.

As manufacturing continues to embrace data-driven approaches, tools that connect industrial data with advanced analytical methods will play an important role. The PyScrew collection and associated library represent a significant step toward more accessible and standardized industrial data resources, facilitating the development of practical solutions for manufacturing quality assurance.

\section{Experimental Setup}
\label{sec:experimental_setup}
The datasets presented in this paper were collected using a standardized experimental setup designed to capture comprehensive data from industrial screw driving operations on the same system. This section details the hardware components, process phases of a typical screw run, and data acquisition methods used in the experiments.

\subsection{Hardware Components}
\begin{figure}[!htbp]
\centering
\includegraphics[width=0.4\textwidth]{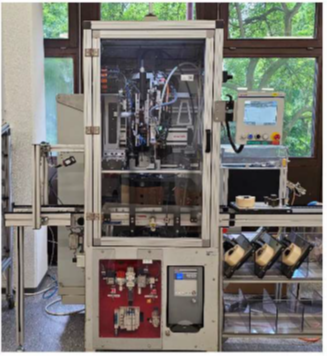}
\caption{The automatic screw driving station used for data collection. The lower section contains pneumatic control with valve islands. The middle section houses the actual screw cell with the linear axis unit and screw tool. On the right are the control panel and screw magazine.}
\label{fig:screw_station}
\end{figure}

The screw driving operations were performed using an automatic screwing station specifically designed for the assembly of EV motor control units (Figure \ref{fig:screw_station}). This band-integrated system represents the state of the art in industrial screw driving technology, combining modern control, sensor, and handling technologies. It is embedded in a fully automated assembly line, representing one of the most complex configurations of screw stations.

The core of the system is a CS351S-D compact system by Bosch Rexroth. The screw spindle configuration consists of:
\begin{itemize}
    \item EC302 servo motor and a 2GE26 planetary gearbox
    \item 2DMC006 measuring transducer for capturing torque, angle, and gradient
    \item Modified 2GB82F73 SZ2 1/4"QC straight output for vacuum screw pickup
\end{itemize}

The process begins with the positioning of two thermoplastic housing parts of a motor control unit, which are designed to prevent rotation or slipping during the screw driving process. For joining the housing parts, two DELTA PT® 40x12 screws from EJOT are used per assembly. These screws are specially designed for thread-forming applications and are characterized by their ability to achieve higher preload forces while simultaneously providing increased breaking torque. The worldwide use of this screw type, especially in the automotive industry, underscores its technological significance.

Screw feeding is accomplished through a elevator rail conveyor equipped with flow control and a controller. A swinging movement of the elevator rail brings the screws into the correct position before they are transported to the screw tool. Workpiece identification occurs at the beginning of the screw process via a Data Matrix Code (DMC) that is applied before the first screw connection and captured by a scanner.

\subsection{Screw Program Phases}
\begin{figure}[!htbp]
\centering
\includegraphics[width=0.8\textwidth]{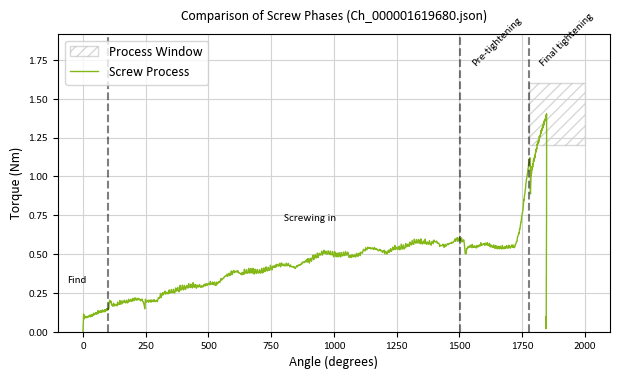}
\caption{Exemplary screw process showing the four characteristic phases: finding, driving in, pre-tightening, and final tightening. The graph displays torque (Nm) versus angle (degrees), with process windows indicated by hatched areas.}
\label{fig:screw_phases}
\end{figure}

The application represents an electrically controlled screw connection (ESV) implemented through a multi-stage screw program. The special feature lies in the use of a self-tapping thread, which places specific demands on process control. As illustrated in Figure \ref{fig:screw_phases}, the process is divided into four characteristic phases, characterized by a combination of torque-controlled and rotation angle-controlled sections.

The first phase, the finding phase (0° to 100°), serves for secure positioning of the screw and is rotation angle-controlled at a lower speed of 150 rpm. Torque monitoring with a maximum value of 1.2 Nm prevents possible damage due to misalignments or jamming of the screw.

In the subsequent phase, the screw-in phase, which is also rotation angle-controlled, the actual thread is formed in the thermoplastic material. This phase is characterized by a characteristic increasing torque profile and extends to a target angle of 1400°. The increased speed of 600 rpm ensures efficient process time while limiting the thermal stress on the plastic. The maximum torque remains limited to 1.2 Nm to avoid overstressing the material.

The transition to the pre-tightening phase marks the change to torque control. With a reduced speed of 200 rpm, a defined target torque of 1.1 Nm is sought. This phase is particularly relevant as it involves the head contact of the screw, and the resulting torque profile provides important information about the quality of the connection.

In the final tightening phase, the speed is greatly reduced to 40 rpm to enable precise adjustment of the target torque of 1.4 Nm. The tight process control with a torque limitation between 1.2 Nm and 1.6 Nm, combined with angle limitation, ensures a reproducible preload force of the connection.

To secure the overall process, indirect screw time monitoring is implemented. The maximum total angle of 2000° serves as an abort criterion for faulty screw processes without restricting the necessary process flexibility during normal variation. This multi-level monitoring strategy enables reliable quality assurance with high process robustness.

\subsection{Data Collection}

For each screw driving operation, the process data acquisition system captures three primary variables:
\begin{itemize}
\item \textbf{Torque [Nm]}: The rotational force applied during the screw driving process
\item \textbf{Angle [degrees]}: The cumulative rotation of the screw during the process
\item \textbf{Gradient [Nm/°]}: The rate of change in torque with respect to angle
\end{itemize}

These measurements are acquired by a Bosch Rexroth BS350 controller with a rotational resolution of 0.25°. Data is sampled at a frequency of 833.33 Hz, resulting in approximately 3-4 data points per degree of rotation. This high sampling rate ensures that even rapid changes in process dynamics are captured accurately.
Data collection begins at the start of the finding phase (0°) and continues until either successful completion of the final tightening phase or abortion due to limit violations. This approach results in variable-length time series, with typical operations generating between 5,000-7,000 data points across all phases.

Each screw driving operation is classified as either successful (OK) or unsuccessful (NOK) based on the following criteria:

\begin{itemize}
\item \textbf{Process completion}: The operation must complete all four phases without triggering abort conditions
\item \textbf{Torque thresholds}: The final torque value must fall within the specified limits (1.2-1.6 Nm)
\item \textbf{Angle limits}: The total rotation angle must not exceed the maximum threshold of 2000°
\item \textbf{Torque gradient profile}: The torque/angle relationship must conform to expected patterns for each phase
\end{itemize}

These criteria ensure that both the process execution and the final connection quality meet the industrial standards for the application.

To ensure robust data collection and integrity, a multi-level architecture was implemented:
\begin{itemize}
\item \textbf{Primary acquisition}: Process data is captured by the BS350 controller in real-time
\item \textbf{Edge processing}: Data is transferred via FTP to an edge device for immediate storage and initial validation
\item \textbf{Data validation}: Automatic checks verify data completeness and sensor readings
\item \textbf{Standardized formatting}: Data is converted to a consistent JSON format with standardized timestamps and measurements
\item \textbf{Archival storage}: Processed data is organized by scenario, class, and operation ID for efficient retrieval
\end{itemize}

This architecture ensured uninterrupted data collection even during temporary network disruptions, maintaining data completeness above 95\% throughout all experiments. The edge device also enabled initial data preprocessing directly at the point of origin, ensuring consistent data quality before central storage and further analysis.

\section{Dataset Description}
\label{sec:dataset_description}
This section describes the six datasets included in the collection. Each dataset focuses on different aspects of the screw driving process, capturing various experimental conditions and scenarios relevant to industrial applications. In total, the collection contains 34,182 individual screw driving operations across six distinct scenarios, representing a comprehensive resource for research in manufacturing quality control and process monitoring.

Table \ref{tab:dataset_overview} provides a summary of the datasets, including their names, number of observations, number of classes, and primary purpose.

\begin{table}[ht]
\caption{Overview of the Screw Driving Datasets.}
\label{tab:dataset_overview}
\centering
\begin{tabular}{p{4.5cm}rcp{6cm}}
\toprule
Scenario name & Obs. & Classes & Purpose \\
\midrule
s01\_variations-in-thread-degradation & 5,000 & 1 & Studies natural degradation of plastic threads over repeated use cycles, documenting wear patterns and failure progression \\
\midrule
s02\_variations-in-surface-friction & 12,500 & 8 & Examines effects of different surface conditions (lubricants, surface treatments, contamination) on screw driving performance \\
\midrule
s03\_variations-in-assembly-conditions-1 & 1,700 & 26 & Investigates diverse component and assembly faults including washer modifications, thread deformations, and alignment issues \\
\midrule
s04\_variations-in-assembly-conditions-2 & 5,000 & 25 & Features methodically arranged assembly fault conditions in 5 distinct error groups with paired normal/abnormal operations \\
\midrule
s05\_variations-in-upper-workpiece-fabrication & 2,400 & 42 & Analyzes how injection molding parameter variations in the upper component affect screw driving metrics \\
\midrule
s06\_variations-in-lower-workpiece-fabrication & 7,482 & 44 & Explores effects of injection molding parameter variations in the lower component on fastening quality \\
\bottomrule
\end{tabular}
\end{table}

All datasets share a common structure and format, providing both raw time series data and associated metadata:
\begin{itemize}
    \item JSON files containing complete measurement data for each screw driving operation
    \item Time series data including torque values, angle values, gradient values, and time values 
    \item Labels CSV file with metadata such as workpiece ID, operation timestamp, usage count, and outcome
    \item Standardized class labels for categorizing experimental conditions
    \item Documentation detailing experimental setup and data interpretation
\end{itemize}

All datasets were collected using the same experimental setup described in Section 3, with a sampling frequency of 833.33 Hz, maintaining data completeness above 95\% throughout the collection. Each time series captures the complete screw driving operation across all four phases: finding, thread forming, pre-tightening, and final tightening. These datasets can be used to develop and test advanced methods, such as anomaly detection \cite{schlegl2021a}, feature extraction techniques \cite{west2021}, or classification algorithms for imbalanced data \cite{schlegl2021b}, which are common challenges in manufacturing process monitoring.

\subsection{Thread Degradation (s01)}
The Thread Degradation dataset (s01\_variations-in-thread-degradation) captures the natural degradation of plastic threads over repeated use. Unlike other datasets in the collection, this scenario focuses exclusively on normal wear patterns without introducing artificial defects, providing a baseline for understanding how thread quality degrades through repeated use.

The dataset includes 5,000 screw driving operations conducted on 100 unique workpieces, with each workpiece subjected to 25 cycles per location across two locations (left/right). This repeated usage creates a natural progression from pristine threads to increasingly worn connections, eventually leading to thread failure in some cases. The dataset shows an overall failure rate of 18.22\%, with failures increasing progressively with usage count.

This dataset is particularly valuable for research in predictive maintenance and early fault detection, as it demonstrates the gradual transition from normal operation to failure under standard conditions. The class structure is shown in Table \ref{tab:s01_classes}.

\begin{table}[ht]
\caption{Classes in the Thread Degradation Dataset (s01)}
\label{tab:s01_classes}
\centering
\begin{tabular}{p{3.5cm}rp{1.5cm}p{6.5cm}}
\toprule
Class name & Samples & Condition & Description \\
\midrule
001\_control-group & 5,000 & normal & No additional manipulations, only wear down from repeated use \\
\bottomrule
\end{tabular}
\end{table}

While containing only a single class, the dataset provides rich information through the workpiece usage count, which ranges from 0 (first use) to 24 (final use). This allows for detailed analysis of degradation patterns over time, with samples naturally distributed across the component lifecycle.

\subsection{Surface Friction (s02)}
The Surface Friction dataset (s02\_variations-in-surface-friction) examines how different surface conditions affect screw driving performance. With 12,500 observations across 8 distinct surface conditions, this dataset represents the largest single scenario in the collection by total observation count. The experiments capture various real-world scenarios including normal use, component wear, lubrication effects, surface treatments, and contamination. This dataset has proven particularly valuable for benchmarking feature extraction methods, as demonstrated by Henkies et al. \cite{henkies2025}, who used it to evaluate the efficiency and effectiveness of different dimensionality reduction approaches for screw connection quality monitoring.

The dataset was collected using 250 unique workpieces, with each workpiece undergoing 25 cycles per location across two locations (left/right). The overall failure rate is 23.90\%, with significant variation between different surface conditions—from as low as 9.08\% with used upper workpieces to as high as 60.96\% with oil-based lubricant contamination.

This dataset is particularly valuable for understanding how surface properties influence the screw driving process, with applications in quality control, contamination detection, and manufacturing process optimization. The class structure is detailed in Table \ref{tab:s02_classes}.

\begin{table}[ht]
\caption{Classes in the Surface Friction Dataset (s02)}
\label{tab:s02_classes}
\centering
\begin{tabular}{p{3.5cm}rp{1.5cm}p{6.5cm}}
\toprule
Class name & Samples & Condition & Description \\
\midrule
001\_control-group & 2,500 & normal & No additional manipulations, baseline data for reference \\
101\_used-upper-workpiece & 2,500 & faulty & Upper workpiece was used 25 times, showing surface weardown \\
201\_lubricant-water & 1,250 & faulty & Decreased friction due to water-contaminated workpiece surface \\
202\_lubricant-oil-based & 1,250 & faulty & Decreased friction due to lubricant-contaminated workpiece surface \\
301\_sanding-coarse & 1,250 & faulty & Increased friction due to coarse surface treatment by sanding (40 grit) \\
302\_sanding-fine & 1,250 & faulty & Increased friction due to fine surface treatment by sanding (400 grit) \\
401\_plastic-adhesive & 1,250 & faulty & Alien material by producing adhesive-contaminated surfaces \\
402\_surface-chipped & 1,250 & faulty & Alien material by a chip due to mechanically damaged surfaces \\
\bottomrule
\end{tabular}
\end{table}

The design of the dataset enables paired comparisons between related error conditions (e.g., water vs. oil contamination, coarse vs. fine sanding), facilitating detailed analysis of how specific surface modifications affect performance metrics.

\subsection{Assembly Conditions 1 (s03)}
The first Assembly Conditions dataset (s03\_variations-in-assembly-conditions-1) examines how various screw and component faults affect screw driving performance. This dataset includes 1,700 observations across 26 experimental conditions, capturing a diverse range of potential issues encountered in manufacturing environments.

The experimental conditions include washer modifications, thread deformations, interface interferences, structural modifications, and alignment issues, providing a comprehensive survey of possible failure modes. The dataset uses 869 unique workpieces, with an average of 1.96 operations per workpiece. The overall failure rate is 37.24\%, the highest in the collection, reflecting the focus on problematic assembly scenarios.

This dataset is valuable for fault diagnosis and classification in screw driving operations, helping researchers develop systems that can identify specific types of assembly issues from process data. The class structure is detailed in Table \ref{tab:s03_classes}.

\begin{table}[!htbp]
\caption{Classes in the Assembly Conditions 1 Dataset (s03)}
\label{tab:s03_classes}
\centering
\resizebox{\textwidth}{!}{%
\begin{tabular}{p{4.8cm}rp{1.2cm}p{7.5cm}}
\toprule
Class name & Samples & Condition & Description \\
\midrule
001\_control-group-1 & 100 & normal & No manipulations, standard reference data recorded in March 2023 \\
002\_control-group-2 & 100 & normal & No manipulations, standard reference data recorded in February 2024 \\
003\_control-group-from-s01 & 200 & normal & No manipulations, using first cycles from s01 control group \\
004\_control-group-from-s02 & 100 & normal & No manipulations, using first cycles from s02 control group \\
\midrule
101\_m4-washer-in-upper-piece & 50 & faulty & Reduced insertion depth using custom 4mm polyamide washer \\
102\_m3-washer-in-upper-piece & 50 & faulty & Reduced insertion depth using standard M3 polyamide washer \\
103\_m3-half-washer-in-upper-part & 50 & faulty & Asymmetric load distribution using half of an M3 washer \\
\midrule
201\_adhesive-thread & 50 & faulty & Metal adhesive on screw tip creating partial binding \\
202\_deformed-thread-1 & 100 & faulty & Damaged thread by mechanical deformation, recorded March 2023 \\
203\_deformed-thread-2 & 50 & faulty & Damaged thread like 202, recorded three months later \\
\midrule
301\_material-in-the-screw-head & 50 & faulty & Melted adhesive on screw head causing driver slippage \\
302\_material-in-the-lower-part & 50 & faulty & Plastic adhesive in the lower workpiece's screw-in hole \\
\midrule
401\_drilling-out-the-workpiece & 50 & faulty & Enlarged pilot holes in lower part with fine drill \\
402\_shortening-the-screw-1 & 50 & faulty & Removed ~2 thread turns and tip, recorded June 2023 \\
403\_shortening-the-screw-2 & 50 & faulty & Like 402, recorded three months later \\
404\_tearing-off-the-screw-1 & 50 & faulty & Partially sawed screw shaft causing complete failure \\
405\_tearing-off-the-screw-2 & 50 & faulty & Like 404, recorded three months later \\
\midrule
501\_offset-of-the-screw-hole & 50 & faulty & Horizontal misalignment using washer to offset screwdriver \\
502\_offset-of-the-work-piece & 50 & faulty & Angular misalignment between screw axis and insertion tube \\
\midrule
601\_surface-used & 100 & faulty & Used upper workpiece, taken from s02-101 \\
602\_surface-moisture & 50 & faulty & Water contamination, taken from s02-201 \\
603\_surface-lubricant & 50 & faulty & Lubricant contamination, taken from s02-202 \\
604\_surface-adhesive & 50 & faulty & Adhesive contamination, taken from s02-401 \\
605\_surface-sanded-40 & 50 & faulty & Coarse surface treatment, taken from s02-301 \\
606\_surface-sanded-400 & 50 & faulty & Fine surface treatment, taken from s02-302 \\
607\_surface-scratched & 50 & faulty & Mechanically damaged surfaces, taken from s02-402 \\
\bottomrule
\end{tabular}%
}
\end{table}

A notable feature of this dataset is the inclusion of multiple control groups collected at different times, as well as repetitions of some error conditions after a three-month interval, allowing for analysis of temporal consistency in error patterns.

\subsection{Assembly Conditions 2 (s04)}
The second Assembly Conditions dataset (s04\_variations-in-assembly-conditions-2) builds on the methodological insights from s03, providing a more structured approach to error classification. This dataset includes 5,000 observations across 25 experimental conditions, organized into five distinct error groups.

A key methodological improvement in s04 is the alternating sequence of normal and faulty operations (5 OK followed by 5 NOK) within each error class. This approach minimizes environmental condition influences and enables direct comparison between normal and faulty conditions within the same experimental context.

The dataset was collected using 2,500 unique workpieces, with each workpiece used for 2 operations. The overall failure rate is 13.86\%, with variation across different fault types. The experimental conditions are organized into five groups: screw quality deviations, contact surface modifications, component/thread modifications, environmental/interface conditions, and process parameter variations.

\begin{table}[ht]
\caption{Classes in the Assembly Conditions 2 Dataset (s04)}
\label{tab:s04_classes}
\centering
\resizebox{\textwidth}{!}{%
\begin{tabular}{p{4.8cm}rp{1.2cm}p{7.5cm}}
\toprule
Class name & Samples & Condition & Description \\
\midrule
001\_control-group & 200 & mixed & No manipulations, standard reference data \\
\midrule
101\_deformed-thread & 200 & mixed & Damaged thread by mechanically deforming lower section \\
102\_filed-screw-tip & 200 & mixed & Removed thread at screw tip on one side through processing \\
103\_glued-screw-tip & 200 & mixed & Metal adhesive on first 3-4mm of screw tip \\
104\_coated-screw & 200 & mixed & Altered surface properties with different screw coating \\
105\_worn-out-screw & 200 & mixed & Screws with significant wear marks from multiple use cycles \\
\midrule
201\_damaged-contact-surface & 200 & mixed & Two symmetrical damages in upper part contact area \\
202\_broken-contact-surface & 200 & mixed & Continuous crack in upper part contact surface \\
203\_metal-ring-upper-part & 200 & mixed & Metallic O-ring in screw head contact area \\
204\_rubber-ring-upper-part & 200 & mixed & Rubber O-ring in contact surface \\
205\_different-material & 200 & mixed & Upper part made from non-standard plastic material \\
\midrule
301\_plastic-pin-screw-hole & 200 & mixed & Plastic pin in screw-in area of lower part \\
302\_enlarged-screw-hole & 200 & mixed & Lower part screw hole enlarged using 4mm drill \\
303\_less-glass-fiber & 200 & mixed & Lower part with reduced glass fiber content (10\% vs 30\%) \\
304\_glued-screw-hole & 200 & mixed & Plastic adhesive on inner surface of screw hole \\
305\_gap-between-parts & 200 & mixed & 1.1mm metal wire between upper and lower parts \\
\midrule
401\_surface-lubricant & 200 & mixed & Multi-purpose oil (WD40) on head contact surface \\
402\_surface-moisture & 200 & mixed & Water particles (0.5ml) applied to contact area \\
403\_plastic-chip & 200 & mixed & Elongated plastic chip in screw hole \\
404\_increased-temperature & 200 & mixed & Components thermally conditioned in oven \\
405\_decreased-temperature & 200 & mixed & Components cooled in insulated box with ice \\
\midrule
501\_increased-ang-velocity & 200 & mixed & Angular velocity increased 10\% across all phases \\
502\_decreased-ang-velocity & 200 & mixed & Angular velocity reduced 10\% in all process phases \\
503\_increased-torque & 200 & mixed & Target tightening torque increased to 1.5 Nm \\
504\_decreased-torque & 200 & mixed & Target tightening torque decreased to 1.3 Nm \\
\bottomrule
\end{tabular}%
}
\end{table}

This dataset is particularly valuable for developing and validating classification algorithms for assembly faults, with applications in automated quality control systems.

\subsection{Upper Workpiece Fabrication (s05)}
The Upper Workpiece Fabrication dataset (s05\_variations-in-upper-workpiece-fabrication) examines how variations in injection molding parameters for the upper workpiece affect screw driving performance. This dataset includes 2,400 observations across 42 experimental conditions, systematically exploring the effects of material composition and manufacturing process parameters.

The dataset focuses on five parameter groups: glass fiber content, recyclate content, switching point, injection velocity, and mold temperature. Each parameter is varied systematically to simulate real-world manufacturing fluctuations and material batch changes.

The dataset was collected using 1,200 unique workpieces, with each workpiece used for 2 operations. Notably, this dataset shows a very high OK rate (99.88\%), indicating that most parameter variations had minimal impact on screw driving success, though they may still affect process characteristics and final connection quality.

\begin{table}[!htbp]
\caption{Classes in the Upper Workpiece Fabrication Dataset (s05)}
\label{tab:s05_classes}
\centering
\resizebox{\textwidth}{!}{%
\begin{tabular}{p{4.8cm}rp{1.2cm}p{7.5cm}}
\toprule
Class name & Samples & Condition & Description \\
\midrule
101\_glass-fiber-content-30 & 80 & normal & Standard material with 30\% glass fiber content \\
102\_glass-fiber-content-28 & 80 & faulty & Reduced glass fiber content to 28\% \\
103\_glass-fiber-content-26 & 80 & faulty & Reduced glass fiber content to 26\% \\
104\_glass-fiber-content-24 & 80 & faulty & Reduced glass fiber content to 24\% \\
105\_glass-fiber-content-22 & 80 & faulty & Reduced glass fiber content to 22\% \\
106\_glass-fiber-content-20 & 80 & faulty & Reduced glass fiber content to 20\% \\
107\_glass-fiber-content-18 & 80 & faulty & Reduced glass fiber content to 18\% \\
\midrule
201\_recyclate-content-000 & 80 & normal & Standard material with 0\% recyclate content \\
202\_recyclate-content-010 & 80 & faulty & Addition of 10\% regrind material \\
203\_recyclate-content-020 & 80 & faulty & Addition of 20\% regrind material \\
204\_recyclate-content-030 & 80 & faulty & Addition of 30\% regrind material \\
205\_recyclate-content-040 & 80 & faulty & Addition of 40\% regrind material \\
206\_recyclate-content-050 & 80 & faulty & Addition of 50\% regrind material \\
207\_recyclate-content-060 & 80 & faulty & Addition of 60\% regrind material \\
208\_recyclate-content-070 & 80 & faulty & Addition of 70\% regrind material \\
209\_recyclate-content-080 & 80 & faulty & Addition of 80\% regrind material \\
210\_recyclate-content-090 & 80 & faulty & Addition of 90\% regrind material \\
211\_recyclate-content-100 & 80 & faulty & Complete 100\% regrind material \\
\midrule
301\_switching-point-15-1 & 40 & normal & First reference with 15 cm³ switching point \\
302\_switching-point-16 & 40 & faulty & Increased switching point to 16 cm³ \\
303\_switching-point-17 & 40 & faulty & Increased switching point to 17 cm³ \\
304\_switching-point-18 & 40 & faulty & Increased switching point to 18 cm³ \\
305\_switching-point-19 & 40 & faulty & Increased switching point to 19 cm³ \\
306\_switching-point-15-2 & 40 & normal & Second reference with 15 cm³ switching point \\
307\_switching-point-14 & 40 & faulty & Decreased switching point to 14 cm³ \\
308\_switching-point-13 & 40 & faulty & Decreased switching point to 13 cm³ \\
309\_switching-point-12 & 40 & faulty & Decreased switching point to 12 cm³ \\
310\_switching-point-11 & 40 & faulty & Decreased switching point to 11 cm³ \\
\midrule
401\_injection-velocity-60-1 & 40 & normal & First reference with 60 cm³/s injection velocity \\
402\_injection-velocity-70 & 40 & faulty & Increased injection velocity to 70 cm³/s \\
403\_injection-velocity-80 & 40 & faulty & Increased injection velocity to 80 cm³/s \\
404\_injection-velocity-90 & 40 & faulty & Increased injection velocity to 90 cm³/s \\
405\_injection-velocity-100 & 40 & faulty & Increased injection velocity to 100 cm³/s \\
406\_injection-velocity-60-2 & 40 & normal & Second reference with 60 cm³/s injection velocity \\
407\_injection-velocity-50 & 40 & faulty & Decreased injection velocity to 50 cm³/s \\
408\_injection-velocity-40 & 40 & faulty & Decreased injection velocity to 40 cm³/s \\
409\_injection-velocity-30 & 40 & faulty & Decreased injection velocity to 30 cm³/s \\
410\_injection-velocity-20 & 40 & faulty & Decreased injection velocity to 20 cm³/s \\
\midrule
501\_mold-temperature-30 & 40 & normal & Standard mold temperature of 30°C \\
502\_mold-temperature-35 & 40 & faulty & Increased mold temperature to 35°C \\
503\_mold-temperature-40 & 40 & faulty & Increased mold temperature to 40°C \\
504\_mold-temperature-45 & 40 & faulty & Increased mold temperature to 45°C \\
\bottomrule
\end{tabular}%
}
\end{table}

This dataset is valuable for understanding the relationship between manufacturing parameters and assembly performance, with applications in process optimization and material engineering.

\subsection{Lower Workpiece Fabrication (s06)}
The Lower Workpiece Fabrication dataset (s06\_variations-in-lower-workpiece-fabrication) complements s05 by focusing on the injection molding parameters for the lower workpiece. With 7,482 observations across 44 experimental conditions, this is the second-largest dataset in the collection, providing comprehensive coverage of manufacturing parameter effects.

Similar to s05, this dataset explores five parameter groups: cooling time, mold temperature, glass fiber content, switching point, and injection velocity. Each parameter is varied systematically across multiple levels to understand threshold effects and process sensitivities.

The dataset was collected using 3,743 unique workpieces, with each workpiece used for 2 operations. The overall failure rate is 2.78\%, with significant variation between parameter groups—notably, reduced glass fiber content shows a strong correlation with increased failure rates, with up to 43.27\% NOK rates at the lowest glass fiber content (10\%).

\begin{table}[!htbp]
\caption{Classes in the Lower Workpiece Fabrication Dataset (s06)}
\label{tab:s06_classes}
\centering
\resizebox{\textwidth}{!}{%
\begin{tabular}{p{4.8cm}rp{1.2cm}p{7.5cm}}
\toprule
Class name & Samples & Condition & Description \\
\midrule
001\_control-group-01 & 240 & normal & Standard reference material and process parameters \\
\midrule
101\_cooling-time-25-1 & 269 & normal & Baseline cooling time of 25 seconds \\
102\_cooling-time-26 & 237 & faulty & Increased cooling time to 26 seconds \\
103\_cooling-time-27 & 232 & faulty & Increased cooling time to 27 seconds \\
104\_cooling-time-28 & 236 & faulty & Increased cooling time to 28 seconds \\
105\_cooling-time-29 & 200 & faulty & Increased cooling time to 29 seconds \\
106\_cooling-time-30 & 156 & faulty & Increased cooling time to 30 seconds \\
107\_cooling-time-25-2 & 190 & normal & Second reference with 25 seconds cooling time \\
\midrule
201\_mold-temperature-30 & 252 & normal & Standard mold temperature of 30°C \\
202\_mold-temperature-35 & 160 & faulty & Increased mold temperature to 35°C \\
203\_mold-temperature-45 & 330 & faulty & Increased mold temperature to 45°C \\
204\_mold-temperature-55 & 248 & faulty & Increased mold temperature to 55°C \\
205\_mold-temperature-65 & 166 & faulty & Increased mold temperature to 65°C \\
\midrule
301\_glass-fiber-content-30 & 326 & normal & Standard material with 30\% glass fiber content \\
302\_glass-fiber-content-25 & 268 & faulty & Reduced glass fiber content to 25\% \\
303\_glass-fiber-content-20 & 274 & faulty & Reduced glass fiber content to 20\% \\
304\_glass-fiber-content-15 & 254 & faulty & Reduced glass fiber content to 15\% \\
305\_glass-fiber-content-10 & 104 & faulty & Reduced glass fiber content to 10\% \\
\midrule
401\_switching-point-22-1 & 352 & normal & First reference with 22 cm³ switching point \\
402\_switching-point-20 & 205 & faulty & Decreased switching point to 20 cm³ \\
403\_switching-point-18 & 100 & faulty & Decreased switching point to 18 cm³ \\
404\_switching-point-17 & 100 & faulty & Decreased switching point to 17 cm³ \\
405\_switching-point-16 & 154 & faulty & Decreased switching point to 16 cm³ \\
406\_switching-point-15 & 98 & faulty & Decreased switching point to 15 cm³ \\
407\_switching-point-14 & 108 & faulty & Decreased switching point to 14 cm³ \\
408\_switching-point-13 & 96 & faulty & Decreased switching point to 13 cm³ \\
409\_switching-point-22-2 & 100 & normal & Second reference with 22 cm³ switching point \\
410\_switching-point-24 & 98 & faulty & Increased switching point to 24 cm³ \\
411\_switching-point-26 & 95 & faulty & Increased switching point to 26 cm³ \\
412\_switching-point-28 & 102 & faulty & Increased switching point to 28 cm³ \\
413\_switching-point-30 & 100 & faulty & Increased switching point to 30 cm³ \\
414\_switching-point-32 & 104 & faulty & Increased switching point to 32 cm³ \\
\midrule
501\_injection-velocity-030-1 & 350 & normal & First reference with 30 cm³/s injection velocity \\
502\_injection-velocity-050 & 100 & faulty & Increased injection velocity to 50 cm³/s \\
503\_injection-velocity-070 & 98 & faulty & Increased injection velocity to 70 cm³/s \\
504\_injection-velocity-090 & 110 & faulty & Increased injection velocity to 90 cm³/s \\
505\_injection-velocity-110 & 98 & faulty & Increased injection velocity to 110 cm³/s \\
506\_injection-velocity-130 & 130 & faulty & Increased injection velocity to 130 cm³/s \\
507\_injection-velocity-150 & 98 & faulty & Increased injection velocity to 150 cm³/s \\
508\_injection-velocity-170 & 102 & faulty & Increased injection velocity to 170 cm³/s \\
509\_injection-velocity-030-2 & 100 & normal & Second reference with 30 cm³/s injection velocity \\
510\_injection-velocity-025 & 138 & faulty & Decreased injection velocity to 25 cm³/s \\
511\_injection-velocity-020 & 106 & faulty & Decreased injection velocity to 20 cm³/s \\
512\_injection-velocity-015 & 98 & faulty & Decreased injection velocity to 15 cm³/s \\
\bottomrule
\end{tabular}%
}
\end{table}

This dataset is valuable for understanding critical thresholds in manufacturing parameters and their impact on assembly quality.

\section{Data Access}
\label{sec:data_access}
The datasets described in this paper are publicly available and can be accessed in two ways: directly through the Zenodo repository for raw data access, or programmatically using the PyScrew Python library for convenient data loading and processing.

\subsection{Accessing the PyScrew Collection via Zenodo}
All six datasets in the PyScrew collection are hosted on Zenodo, a general-purpose open repository that allows researchers to deposit datasets, software, reports, and other research-related artifacts. The datasets are accessible through the persistent DOI: \url{https://doi.org/10.5281/zenodo.14729547}, which always points to the latest version of the collection (currently v1.2.2 as of May 13, 2025).

Each dataset is provided as a separate ZIP file, with standardized organization:
\begin{itemize}
    \item Complete raw data in JSON format organized in subdirectories per class
    \item Standardized labels.csv files for metadata and classification
    \item Comprehensive README.md documentation for each scenario
\end{itemize}

The collection on Zenodo includes the following files:
\begin{itemize}
    \item s01\_variations-in-thread-degradation.zip (29.0 MB)
    \item s02\_variations-in-surface-friction.zip (83.9 MB)
    \item s03\_variations-in-assembly-conditions-1.zip (13.2 MB)
    \item s04\_variations-in-assembly-conditions-2.zip (61.1 MB)
    \item s05\_variations-in-upper-workpiece-fabrication.zip (26.7 MB)
    \item s06\_variations-in-lower-workpiece-fabrication.zip (83.1 MB)
\end{itemize}

When using these datasets in academic work, researchers are requested to cite the dataset using the following citation:

West, N., \& Deuse, J. (2025). Industrial screw driving dataset collection: Time series data for process monitoring and anomaly detection (v1.2.1) [Data set]. Nikolai West. \url{https://doi.org/10.5281/zenodo.15273503}

\subsection{Accessing PyScrew through the Python Library}
The PyScrew library provides a more convenient way to access and work with the datasets, especially for Python users. Available on PyPI (\url{https://pypi.org/project/pyscrew/}), the library can be easily installed via pip and handles data downloading, validation, and preprocessing, offering a consistent interface across all datasets. The library is designed to integrate seamlessly with common data science workflows and machine learning frameworks.

PyScrew offers several advantages over direct access to the raw data. It provides automatic downloading and caching, with datasets downloaded on demand and stored locally for future use. The library supports multiple dataset identification methods, allowing scenarios to be referenced by short ID (e.g., "s01"), full name (e.g., "thread-degradation"), or full ID (e.g., "s01\_thread-degradation"). The consistent data format standardizes structure across all scenarios, simplifying analysis. Comprehensive data validation ensures integrity checks for reliability, while the flexible preprocessing pipeline offers options for handling duplicates, missing values, and normalization. The library also includes detailed logging and statistics for transparent processing information.

Loading data with PyScrew requires minimal code. The following example demonstrates how to list available scenarios and load data from the thread degradation dataset with default settings:

\begin{verbatim}
import pyscrew

# List available scenarios
scenarios = pyscrew.list_scenarios()
print("Available scenarios:", scenarios)

# Load data from the thread degradation scenario with default settings
data = pyscrew.get_data(scenario="s01")

# Access the measurements and labels
x_values = data["torque values"]
y_values = data["class values"]
\end{verbatim}

For more specialized research needs, PyScrew offers extensive configuration options to tailor data processing. Researchers can specify which classes to include, which measurement types to return, which screw phases to focus on, and how to handle data preprocessing challenges such as duplicate time points, missing values, and time series length normalization. The following example demonstrates these configuration capabilities:

\begin{verbatim}
data = pyscrew.get_data(
    scenario="s04",                           # Selected dataset, e.g. "s04" for assembly conditions 2
    cache_dir="~/.cache/pyscrew",             # Custom cache directory
    force_download=True,                      # Force re-download even if cached
    scenario_classes=["001_control-group", "101_deformed-thread"],  # Filter by class
    return_measurements=["torque", "angle"],  # Select specific measurements
    screw_phase=[3, 4],                       # Include only phases 3 and 4
    handle_duplicates="first",                # How to handle duplicate time points
    handle_missings="mean",                   # How to handle missing values
    target_length=1000,                       # Target length for normalization
)
\end{verbatim}

This flexibility allows researchers to focus on their specific research questions without spending time on data preparation and processing, which is particularly valuable for rapid prototyping and experimentation with different analysis approaches. For example, Henkies et al. \cite{henkies2025} demonstrated how different feature extraction methods can be applied to screw connection data using the dataset s02, with each approach offering distinct advantages depending on the computational resources available and the specific classification goals. Another example is the work from West et al. \cite{west2025} that explores the capability to detect specific error classes among a large group of distinct errors using the dataset s04. Their work highlights the importance of having standardized, accessible datasets that allow for methodological comparisons.

PyScrew is designed to work seamlessly with popular machine learning libraries, making it straightforward to incorporate the screw driving datasets into existing analysis workflows. The following example demonstrates how to use PyScrew with scikit-learn for a basic classification task using the surface friction dataset:

\begin{verbatim}
import pyscrew
from sklearn.model_selection import train_test_split
from sklearn.ensemble import RandomForestClassifier
from sklearn.metrics import classification_report

# Load data from surface friction dataset
data = pyscrew.get_data(scenario="s02")

# Prepare features and target
X = data["torque_values"]
y = data["class_values"]

# Split data
X_train, X_test, y_train, y_test = train_test_split(
    X, y, test_size=0.3, random_state=42
)

# Train a classifier
clf = RandomForestClassifier(n_estimators=100)
clf.fit(X_train, y_train)

# Evaluate performance
y_pred = clf.predict(X_test)
print(classification_report(y_test, y_pred))
\end{verbatim}

PyScrew follows a modular design with comprehensive documentation. Detailed information about each dataset is available both in the GitHub repository (\url{https://github.com/nikolaiwest/pyscrew}) and through the package's listing functionality. The package structure includes core data models, configuration management, processing pipeline components, and scenario-specific configurations, all organized for maintainability and extensibility. The library is actively maintained and includes extensive tests to ensure reliability.

The default cache structure stores downloaded data in a user-specific cache directory (\texttt{\textasciitilde/.cache/pyscrew/}), with separate subdirectories for compressed archives and extracted dataset files. This approach minimizes redundant downloads and optimizes disk usage while ensuring data availability for offline use.

While PyScrew is the recommended method for accessing and processing the datasets, all data remains available in raw format through Zenodo for researchers who prefer to implement custom processing pipelines or work in programming languages other than Python. This dual-access approach maximizes the datasets' accessibility and utility for the broader research community.

\section{Limitations and Future Work}
\label{sec:limitations}
PyScrew addresses a significant gap in the accessibility of industrial manufacturing data for research purposes. By providing a consistent interface to comprehensive screw driving datasets with detailed documentation, the package enables researchers to develop and validate methodologies for quality control, anomaly detection, and process monitoring. Future work could explore the integration of interpretable deep learning approaches for anomaly detection \cite{schlegl2021a}, efficient feature extraction methods \cite{west2021}, and robust classification techniques for imbalanced data \cite{schlegl2021b} with the PyScrew datasets. The flexible processing pipeline accommodates diverse research requirements while ensuring reproducibility through standardized data handling.

Several promising directions for future development of PyScrew include expanding the collection with additional datasets covering other aspects of screw driving processes, implementing specialized signal processing methods for feature extraction, providing reference implementations of common machine learning models, extending the package to support real-time processing, and including complementary datasets from related manufacturing processes. The modular design of PyScrew facilitates these extensions while maintaining compatibility with existing applications.

While PyScrew provides a valuable resource for research in manufacturing quality control, several limitations should be acknowledged. The datasets were collected in a controlled laboratory setting, which may not fully represent the variability encountered in industrial production environments. The focus on a specific type of plastic component and screw limits generalization to other material combinations. The 833.33 Hz sampling frequency, while high, may not capture ultra-high-frequency phenomena that could be relevant for certain fault types. Despite these limitations, PyScrew represents a significant step toward more accessible and standardized industrial data resources.

As manufacturing continues to embrace data-driven approaches, tools like PyScrew will play an important role in connecting industrial data with advanced analytical methods. We hope this contribution will accelerate research in this domain and promote the development of practical solutions for manufacturing quality assurance.

\section{Conclusion}
\label{sec:conclusion}
This paper has presented a comprehensive collection of six datasets from industrial screw driving experiments, covering various aspects including thread degradation, surface friction, assembly conditions, and workpiece fabrication parameters. Together, these datasets comprise over 34,000 individual screw driving operations, providing a substantial resource for researchers working on industrial process monitoring, quality control, anomaly detection, and machine learning applications in manufacturing.

The key contributions of this work include:

\begin{itemize}
    \item A standardized collection of industrial screw driving datasets with detailed documentation, capturing diverse experimental conditions relevant to manufacturing applications
    \item Comprehensive experimental design that systematically explores factors affecting screw driving processes in plastic components
    \item Open access to both raw data (via \href{https://doi.org/10.5281/zenodo.14729547}{Zenodo}) and processed data (via the \href{https://github.com/nikolaiwest/pyscrew}{PyScrew library})    \item A flexible data access framework that facilitates reproducible research and fair comparison of methodologies
\end{itemize}

Each dataset in the collection is designed to address specific research questions in manufacturing quality control. The thread degradation dataset (s01) provides insights into natural wear patterns, while the surface friction dataset (s02) examines the effects of different surface conditions. The assembly conditions datasets (s03 and s04) focus on various fault scenarios, with s04 offering a methodologically improved design. The workpiece fabrication datasets (s05 and s06) investigate how manufacturing parameters for different components affect the screw driving process.

By making these datasets publicly available, we aim to address the scarcity of standardized industrial datasets in the research community. The datasets enable researchers to develop and validate new methodologies for industrial quality control, process monitoring, and fault diagnosis without the need to set up extensive experimental infrastructure.

The PyScrew library further enhances the accessibility of these datasets by providing a consistent interface for data loading, preprocessing, and integration with common machine learning workflows. This approach promotes reproducibility and standardization in research on industrial screw driving processes.

We believe that these resources will contribute to advancing research in industrial data analysis and facilitate the development of practical solutions for manufacturing quality assurance. As manufacturing continues to embrace data-driven approaches, standardized datasets like those presented in this paper will play an increasingly important role in connecting industrial data with advanced analytical methods.

\section*{Acknowledgments}
These datasets were collected and prepared by the RIF Institute for Research and Transfer e.V. and the Technical University Dortmund, Institute for Production Systems. The research was supported by the German Ministry of Education and Research (BMBF) and the European Union's "NextGenerationEU" program as part of the funding program "Data competencies for early career researchers."

%Bibliography
\bibliographystyle{unsrt}

\end{document}